# $\beta$-Divergence-Based Latent Factorization of Tensors model for QoS prediction

Zemiao Peng, Hao Wu

*Abstract*—A non-negative latent factorization of tensors (NLFT) model can well model the temporal pattern hidden in non-negative quality-of-service (QoS) data for predicting the unobserved ones with high accuracy. However, existing NLFT models' objective function is based on Euclidean distance, which is only a special case of $\beta$-divergence. Hence, can we build a generalized NLFT model via adopting $\beta$-divergence to achieve prediction accuracy gain? To tackle this issue, this paper proposes a $\beta$-divergence-based NLFT model ($\beta$-NLFT). Its ideas are two-fold: 1) building a learning objective with $\beta$-divergence to achieve higher prediction accuracy; and 2) implementing self-adaptation of hyper-parameters to improve practicability. Empirical studies on two dynamic QoS datasets demonstrate that compared with state-of-the-art models, the proposed $\beta$-NLFT model achieves the higher prediction accuracy for unobserved QoS data.

## I. Introduction

WITH RAPID GROWTH OF THE NUMBER OF WEB SERVICES, many functionally similar or identical Web services are provided by different providers [1]. Thus, a designer of service-oriented application needs to choose suitable Web services for creating high-quality applications [2]-[4]. In particular, the non-functional quality-of-service (QoS) is a major indicator to judge the performance of a Web service, where user-perspective QoS, e.g., response time and throughput, provides more direct clues to help a designer select suitable Web services [5]. However, due to some practical limitations like drastically increasing service count and high invoking expenses, only a small part of user-perspective QoS data can be observed, thus, the available QoS data are often high dimensional and incomplete (HDI) [6]. Therefore, how to implement accurate prediction for unobserved QoS data becomes an essential yet thorny issue.

Over the last few decades, various methods have been presented to tackle this issue. Zheng *et al.* [7] proposes a collaborative QoS prediction method for Web services using the past Web service usage experience of service users. Luo *et al.* [8] proposes a non-negative matrix decomposition model based on regularized single elements, which is particularly suitable for solving collaborative problems subject to non-negativity constraints. Tang *et al.* [9] design a method to integrate network map in matrix decomposition models, which improves the prediction accuracy of traditional matrix decomposition models. Although the aforementioned methods can predict unobserved QoS data, they focus on static QoS data and fail to capture dynamic QoS data temporal pattern [10], [12], [16].

In industrial applications, since the service statues and network environment often vary over time, QoS data from the identical Web service can change in real time. Therefore, it becomes crucial to build a sophisticated QoS-predictor that can model temporal pattern of QoS data. As previous research [11]-[13], the dynamic QoS data can be seamlessly represented by an HDI tensor, and its temporal and spatial patterns can be fully reserved [14], [15]. A non-negative latent factorization of tensors (NLFT)-based QoS-predictor can possess highly efficient in predicting unobserved dynamic QoS data [10], [14], [17], [18]. For example, Wu *et al.* [10] build several regularized NLFT models via using different regularization methods. Luo *et al.* [14] incorporate the linear biases into an NLFT model to boost the prediction performance. Zhang *et al.* [18] adopt non-negative multiplicative update rule as model optimization algorithm.

Note that the objective function of the NLFT-based QoS predictors rely on the Euclidean distance, which is only a special case of $\beta = 2$ in $\beta$-divergence [19]. In particular, the objective function of an NLFT model has vital effects to its performance like prediction accuracy. As revealed in [20], $\beta$-divergence is far more viable and flexible than the Euclidean distance in building missing data prediction model. Therefore, is it possible to design an NLFT model via adopting the $\beta$-divergence as objective function, thereby improving its prediction accuracy for missing dynamic QoS data?

Aim at addressing above issue, in this paper we propose a $\beta$-divergence-based NLFT ($\beta$-NLFT) model with a $\beta$-divergence function as its objective function, and implement the hyper-parameters self-adaptation by using particle swarm optimization (PSO) method [30]. The main contributions include:1) A $\beta$-NLFT model. It adopts $\beta$-divergence to build learning objective for achieving higher prediction accuracy; and 2) An adaptive optimization parameters update rule. It implements hyper-parameters self-adaptation by incorporating a PSO algorithm into a single latent factor-dependent, non-negative and multiplicative update on the tensor (SLF-NMUT) algorithm [21], [22].

## II. Preliminaries

*A. QoS Data Tensor*

As shown in Fig. 1, the dynamic QoS data involved in this paper are described by an HDI tensor **Y**, which is defined as follows:

**Definition 1:** Given a QoS data tensor $\mathbf{Y}^{|I|\times|J|\times|K|} \in \mathbb{R}$, its each element $y_{ijk}$ indicates the QoS experienced by user $i \in I$ on service $j \in J$ at time slot $k \in K$. Let $\Lambda$ and $\Gamma$ respectively indicate the known and unknown element sets of **Y**, if $|\Lambda| \ll |\Gamma|$, then **Y** is a HDI tensor. Note that the HDI tensor **Y** is a non-negative tensor because the QoS data produced by the real applications is non-negative [23].

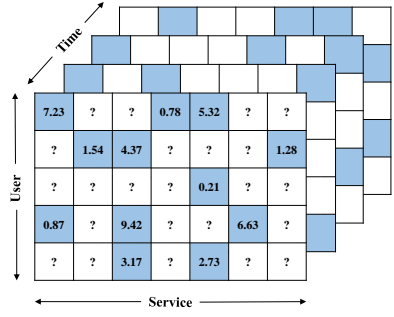

Fig. 1. An HDI user-service-time tensor.

*B. Non-negative Latent Factorization of tensors (NLFT)*

Given a non-negative HDI tensor $\mathbf{Y}^{|I|\times|J|\times|K|}$, an NLFT model builds the rank-$R$ approximation $\hat{\mathbf{Y}}^{|I|\times|J|\times|K|}$ of **Y** based on Canonical Polyadic tensor factorization (CPTF) framework [10], [24], where $\hat{\mathbf{Y}}$ consists of $R$ rank-one tensors $\mathbf{A}_r^{|I|\times|J|\times|K|}$ ($r=1,…,R$). Thus, we define the rank-one tensor as follows:

***Definition 2:*** Let $\mathbf{A}_r^{|I|\times|J|\times|K|}$ to be a rank-one tensor, it can be obtained by the outer product of three latent feature (LF) vectors of $u_r$, $s_r$ and $t_r$ with length of $|I|$, $|J|$, and $|K|$, i.e., $A_r = \boldsymbol{u}_r \circ \boldsymbol{s}_r \circ \boldsymbol{t}_r$ [11].

According to Definition 2, we can obtain the detail representation of each element $a_{ijk}$ in $A_r$ as:

$$a_{ijk} = u_{ir} s_{jr} t_{kr}. \tag{1}$$

Correspondingly, each element of $\hat{y}_{ijk}$ is given as:

$$\hat{y}_{ijk} = \sum_{r=1}^{R} a_{ijk} = \sum_{r=1}^{R} u_{ir} s_{jr} t_{kr}. \tag{2}$$

where $u_{ir}$, $s_{jr}$, and $t_{kr}$ respectively denote an element in LF matrices $U^{|I|\times|R|}$, $S^{|J|\times|R|}$ and $T^{|K|\times|R|}$, the three LF matrices consist of $R$ LF vectors [24].

To obtain the LF matrices, an NLFT model build a learning objective via using commonly-adopted Euclidean distance to distinguish **Y** and $\hat{\mathbf{Y}}$, which is defined on $\Lambda$ only [26]. Thus, it is formulated as:

$$\varepsilon = \sum_{y_{ijk} \in \Lambda} \left(y_{ijk} - \hat{y}_{ijk}\right)^2 = \sum_{y_{ijk} \in \Lambda} \left(y_{ijk} - \sum_{r=1}^{R} u_{ir} s_{jr} t_{kr}\right)^2, \tag{3}$$

s.t. $\forall i \in I, \forall j \in J, \forall k \in K, r \in \{1,…,R\}: u_{ir} \geq 0, s_{jr} \geq 0, t_{kr} \geq 0.$

Moreover, as discussed in [11], it is necessary to integrate an appropriate regularization scheme into learning objective for generalizing the model [25], thus, an objective function adopting a $L_2$ regularization can be obtained as follows:

$$\varepsilon = \sum_{y_{ijk} \in \Lambda} \left(\left(y_{ijk} - \hat{y}_{ijk}\right)^2 + \lambda \sum_{r=1}^{R}\left(u_{ir}^2 + s_{jr}^2 + t_{kr}^2\right)\right), \tag{4}$$

s.t. $\forall i \in I, \forall j \in J, \forall k \in K, r \in \{1,…,R\}: u_{ir} \geq 0, s_{jr} \geq 0, t_{kr} \geq 0.$

Further, as shown in [11]-[14], the performance of NLFT model can be enhanced by introducing three linear bias vectors $\boldsymbol{a}^{|I|} \in \mathbb{R}$, $\boldsymbol{b}^{|J|} \in \mathbb{R}$ and $\boldsymbol{c}^{|K|} \in \mathbb{R}$. Hence, by incorporating such three bias vectors into $\varepsilon$, (4) is reformulated as:

$$\varepsilon = \sum_{y_{ijk} \in \Lambda} \left(\left(y_{ijk} - \sum_{r=1}^{R} u_{ir} s_{jr} t_{kr} - a_i - b_j - c_k\right)^2 + \lambda \sum_{r=1}^{R}\left(u_{ir}^2 + s_{jr}^2 + t_{kr}^2\right) + \lambda_b \left(a_i + b_j + c_k\right)\right), \tag{5}$$

s.t. $\forall i \in I, j \in J, k \in K, r \in \{1,2,...,R\}: u_{ir} \geq 0, s_{jr} \geq 0, t_{kr} \geq 0, a_i \geq 0, b_j \geq 0, c_k \geq 0.$

An SLF-NMUT algorithm has proven to be efficient in optimizing NLFT model [20], [21]. Hence, an NLFT model applies it to (5) to achieve the desired learning rules.

## III. A β-NLFT Model

*A. Objective Function*

From previous research [20], [27], we can know that Euclidean distance is a special case of β-divergence. Hence, the β-divergence between **Y** and $\hat{\mathbf{Y}}$ based on $\Lambda$ for an NLFT model is reformulated as follows:

$$\begin{cases} \beta=0 : d_0 = \sum_{y_{ijk}\in\Lambda}\left(\frac{y_{ijk}}{\hat{y}_{ijk}}-\log\frac{y_{ijk}}{\hat{y}_{ijk}}-1\right), \\ \beta=1 : d_1 = \sum_{y_{ijk}\in\Lambda}\left(y_{ijk}\log\frac{y_{ijk}}{\hat{y}_{ijk}}-y_{ijk}+\hat{y}_{ijk}\right), \\ \beta\neq 0 \text{ or } 1 : d_\beta = \sum_{y_{ijk}\in\Lambda}\frac{\left(y_{ijk}^\beta+(\beta-1)\hat{y}_{ijk}^\beta-\beta y_{ijk}\hat{y}_{ijk}^{\beta-1}\right)}{\beta(\beta-1)}. \end{cases} \quad (6)$$

Correspondingly, by incorporating the regularization term and linear biases into (6), the objective function of the $\beta$-NLFT model is given as:

$$\begin{cases} \varepsilon_0 = \sum_{y_{ijk}\in\Lambda}\left(\left(\frac{y_{ijk}}{\hat{y}_{ijk}}-\log\frac{y_{ijk}}{\hat{y}_{ijk}}-1\right)+\lambda\sum_{r=1}^R\left(u_{ir}^2+s_{jr}^2+t_{kr}^2\right)+\lambda_b\left(a_i^2+b_j^2+c_k^2\right)\right), \\ \varepsilon_1 = \sum_{y_{ijk}\in\Lambda}\left(\left(y_{ijk}\log\frac{y_{ijk}}{\hat{y}_{ijk}}-y_{ijk}+\hat{y}_{ijk}\right)+\lambda\sum_{r=1}^R\left(u_{ir}^2+s_{jr}^2+t_{kr}^2\right)+\lambda_b\left(a_i^2+b_j^2+c_k^2\right)\right), \\ \varepsilon_\beta = \sum_{y_{ijk}\in\Lambda}\left(\frac{\left(y_{ijk}^\beta+(\beta-1)\hat{y}_{ijk}^\beta-\beta y_{ijk}\hat{y}_{ijk}^{\beta-1}\right)}{\beta(\beta-1)}+\lambda\sum_{r=1}^R\left(u_{ir}^2+s_{jr}^2+t_{kr}^2\right)+\lambda_b\left(a_i^2+b_j^2+c_k^2\right)\right). \end{cases} \quad (7)$$

$$s.t. \ \forall i\in I, j\in J, k\in K, r\in\{1,2,...,R\}: u_{ir}\geq 0, s_{jr}\geq 0, t_{kr}\geq 0, a_i\geq 0, b_j\geq 0, c_k\geq 0.$$

where $\varepsilon_0$, $\varepsilon_1$ and $\varepsilon_\beta$ denote objective functions when $\beta=0$, $\beta=1$, and $\beta\notin\{0,1\}$, respectively.

### B. Parameter Learning Scheme

To obtain the $\beta$-SLF-NMUT scheme for the $\beta$-NLFT model, we first analysis Karush-Kuhn-Tucker (KKT) conditions for objective functions [20], [21]. Note that for conciseness, we only show the $\beta$-NLF-NMUT scheme for U and $a$ with $\beta\notin\{0,1\}$

Let $\Gamma^{|I|\times R}$, $K^{|J|\times R}$, $L^{|K|\times R}$, $m^{|I|}$, $n^{|J|}$ and $o^{|K|}$ be the Lagrange multipliers corresponding to the constraints U≥0, S≥0, T≥0, $a$≥0, $b$≥0 and $c$≥0 [19], [20], respectively. Then, we obtain the Lagrangian function for (7) as follows:

$$L_\beta = \varepsilon_\beta + tr(\Gamma U^T) + tr(KS^T) + tr(LT^T) + tr(ma^T) + tr(nb^T) + tr(oc^T)$$
$$= \varepsilon_\beta + \sum_i\sum_r \gamma_{ir}u_{ir} + \sum_j\sum_r k_{jr}s_{jr} + \sum_k\sum_r l_{kr}t_{kr} + \sum_i m_i a_i + \sum_j n_j b_j + \sum_k o_k c_k. \quad (8)$$

Then the partial derivatives of $L_\beta$ in (8) regarding to $u_{ir}$, and $a_i$ are obtained as following:

$$\begin{cases} \frac{\partial L_\beta}{\partial u_{ir}} = \frac{\partial \varepsilon_\beta}{\partial u_{ir}}+\gamma_{ir} = \sum_{y_{ijk}\in\Lambda(i)}\left(s_{jr}t_{kr}\hat{y}_{ijk}^{\beta-1}-s_{jr}t_{kr}y_{ijk}\hat{y}_{ijk}^{\beta-2}+2\lambda u_{ir}\right)+\gamma_{ir}, \\ \frac{\partial l_\beta}{\partial a_i} = \frac{\partial \varepsilon_\beta}{\partial a_i}+m_i = \sum_{y_{ijk}\in\Lambda(i)}\left(\hat{y}_{ijk}^{\beta-1}-y_{ijk}\hat{y}_{ijk}^{\beta-2}+2\lambda_b a_i\right)+m_i. \end{cases} \quad (9)$$

The following observations can be summarized by combining (8) and (9):

*a)* We need to set the partial derivatives of $u_{ir}$ and $a_i$ to zero for obtaining the local optimal value of $L_\beta$, respectively.
*b)* For formula (9) the KKT condition is, $\forall i\in I, r\in\{1,..,R\}$: $\gamma_{ir}u_{ir}=0$, $m_i a_i=0$.

According to *a)* and *b)*, the following deduction can be obtained:

$$\begin{cases} u_{ir}\sum_{y_{ijk}\in\Lambda(i)}\left(s_{jr}t_{kr}y_{ijk}\hat{y}_{ijk}^{\beta-2}-s_{jr}t_{kr}\hat{y}_{ijk}^{\beta-1}+2\lambda u_{ir}\right)=0, \\ a_i\sum_{y_{ijk}\in\Lambda(i)}\left(\hat{y}_{ijk}^{\beta-1}-y_{ijk}\hat{y}_{ijk}^{\beta-2}+2\lambda_b a_i\right)=0. \end{cases} \quad (10)$$

Note that (10) can be reformulated as follows:

$$\begin{cases} u_{ir}\sum_{y_{ijk}\in\Lambda(i)}s_{jr}t_{kr}y_{ijk}\hat{y}_{ijk}^{\beta-2} = u_{ir}\sum_{y_{ijk}\in\Lambda(i)}\left(s_{jr}t_{kr}\hat{y}_{ijk}^{\beta-1}+2\lambda u_{ir}\right), \\ a_i\sum_{y_{ijk}\in\Lambda(i)}y_{ijk}\hat{y}_{ijk}^{\beta-2} = a_i\sum_{y_{ijk}\in\Lambda(i)}\left(\hat{y}_{ijk}^{\beta-1}+2\lambda_b a_i\right). \end{cases} \quad (11)$$

Finally, from (11), we get the update rule for $u_{ir}$, and $a_i$:

$$\begin{cases} u_{ir} = u_{ir}\left(\sum_{y_{ijk}\in\Lambda(i)}s_{jr}t_{kr}y_{ijk}\hat{y}_{ijk}^{\beta-2}\bigg/\left(\sum_{y_{ijk}\in\Lambda(i)}s_{jr}t_{kr}\hat{y}_{ijk}^{\beta-1}+\lambda|\Lambda(i)|u_{ir}\right)\right), \\ a_i = a_i\left(\sum_{y_{ijk}\in\Lambda(i)}y_{ijk}\hat{y}_{ijk}^{\beta-2}\bigg/\left(\sum_{y_{ijk}\in\Lambda(i)}\hat{y}_{ijk}^{\beta-1}+\lambda_b|\Lambda(i)|a_i\right)\right). \end{cases} \quad (12)$$

which is the β-SLF-NMUT rule for $u_{ir}$, and $a_i$ with $\beta \notin \{0, 1\}$ in β-NLFT. We can get similar update for S, T, **b** and **c** by similar deduction. When $\beta = 0$ and $\beta = 1$, the β-SLF-NMUT update rule for each latent feature can be obtained in same way.

*C. Hyper-parameters Self-adaptation*

From (12) we can see that there are three hyper-parameters $\beta$, $\lambda$ and $\lambda_b$. If three-fold grid-search is used to obtain the optimal hyper-parameters, it will take a long time, and the model practicability will also be weakened. Therefore, this study uses the PSO algorithm to make them self-adaptation [24]-[26]. In order to achieve the appeal goal, we set up $Q$ particles, and each particle has three dimensions. The three dimensions of particle $q$ represent three hyper-parameters $\beta_q$, $\lambda_q$ and $\lambda_{bq}$. Then each particles position $x_q$ and velocity $v_q$ are defined as:

$$\begin{cases} x_q = [\beta_q, \lambda_q, \lambda_{bq}], \\ v_q = [v_{q\beta}, v_{q\lambda}, v_{q\lambda_b}]. \end{cases} \quad (13)$$

According to the rule of the PSO the evolution of $\beta_q$, $\lambda_q$ and $\lambda_{bq}$ at the *n*-th iteration are given as:

$\forall q \in \{1,...,Q\}$:

$$\begin{cases} v_q^n = \omega v_q^{n-1} + c_1 r_1(\gamma_q^{n-1} - x_q^{n-1}) + c_2 r_2(\hat{\gamma}^{n-1} - x_q^{n-1}), \\ x_q^n = x_q^{n-1} + v_q^n. \end{cases} \quad (14)$$

where $\gamma_q$ and $\hat{\gamma}$ represent that the best position of particle $q$ and the group best position, respectively. For better converge result the position and velocity of each particle must be limited in a certain range [20]:

$$\begin{cases} \hat{x} = [\hat{\beta}, \hat{\lambda}, \hat{\lambda}_b], \check{x} = [\check{\beta}, \check{\lambda}, \check{\lambda}_b], \\ \hat{v} = [\hat{v}_\beta, \hat{v}_\lambda, \hat{v}_{\lambda_b}], \check{v} = [\check{v}_\beta, \check{v}_\lambda, \check{v}_{\lambda_b}]. \end{cases} \quad (15)$$

From [28]-[30], we have $\hat{v} = 0.2 \times (\hat{x} - \check{x})$ and $\check{v} = -\hat{v}$. In order to make the whole swarm fit better in Λ, the following fitness function is adopted:

$$F = \sqrt{\sum_{y_{ijk} \in \phi} (y_{ijk} - \hat{y}_{ijk})^2 / |\phi|}. \quad (16)$$

According to (16), at the *n*-th iteration, $\gamma_q$ and $\hat{\gamma}$ are updated as:

$$\begin{cases} \gamma_q^n = \begin{cases} x_q^{n-1}, F_q^n \leq F_q^{n-1}, \\ \gamma_q^{n-1}, F_q^n > F_q^{n-1}; \end{cases} \\ \hat{\gamma}^n = \begin{cases} x^{n-1}, F_q^n \leq F_q^{n-1}, \\ \hat{\gamma}^{n-1}, F_q^n > F_q^{n-1}. \end{cases} \end{cases} \quad (17)$$

where $F_q^n$ calculates the fitness function value of particle $q$ in the *n*-th iteration.

Note that $\forall q \in \{1,...,Q\}$, $x_q$ is associated with the identical group hyper-parameters [25]. We can know that each iteration is composed of $Q$ sub-iterations, in the $q$ sub-iterations of the *n*-th iteration the parameters update rule is given as follow:

$$\begin{cases} u_{ir}^{n,q} = u_{ir}^{n-1,q} \left( \sum_{y_{ijk} \in \Lambda(i)} s_{jr} t_{kr} y_{ijk} \hat{y}_{ijk}^{x_{q,1}-2} \middle/ \left( \sum_{y_{ijk} \in \Lambda(i)} s_{jr} t_{kr} \hat{y}_{ijk}^{x_{q,1}-1} + x_{q,2} |\Lambda(i)| u_{ir} \right) \right), \\ a_i^{n,q} = a_i^{n-1,q} \left( \sum_{y_{ijk} \in \Lambda(i)} y_{ijk} \hat{y}_{ijk}^{x_{q,1}-2} \middle/ \left( \sum_{y_{ijk} \in \Lambda(i)} \hat{y}_{ijk}^{x_{q,1}-1} + x_{q,3} |\Lambda(i)| a_i \right) \right). \end{cases} \quad (18)$$

where the $x_{q,1}$, $x_{q,2}$ and $x_{q,3}$ respectively represent the value hyper-parameter of $\beta$, $\lambda$ and $\lambda_b$ in the $q$-th sub-iteration. According to [25], the PSO parameters are set to $Q$=20, c1=c2=2, and ω=0.726.

## IV. EXPERIMENTAL RESULTS AND DISCUSSION

*A. General Settings*

*1) Datasets:* Two dynamic QoS datasets collected by WSMonitor [15] are adopted to validate the performance of the proposed β-NLFT model, and the details of these two databases are summarized in Table I. In our experiments, we randomly divide each dataset into training set ψ, validation set ϕ and testing set φ with the ratio of 7:1:2.

TABLE I. Datasets detials

| Dataset | D1 | D2 |
| --- | --- | --- |
| Data Type | Response-Time | Throughput |
| Scale | 0-20s | 0-1,000kbps |
| User Count | 142 | 142 |

| Service Count | 4,532 | 4,532 |
| Time Point Count | 64 | 64 |
| Element Count | 30,287,611 | 30,287,611 |

*2) Evaluation Metrics:* To measure the prediction accuracy of a tested model, the root mean square error (RMSE) and mean absolute error (MAE) are adopted as metrics [31]-[34], they are defined as:

$$\text{MAE} = \sum_{y_{ijk} \in \varphi} \left| y_{ijk} - \hat{y}_{ijk} \right| / |\varphi|,$$

$$\text{RMSE} = \sqrt{\sum_{y_{ijk} \in \varphi} (y_{ijk} - \hat{y}_{ijk})^2 / |\varphi|}.$$

For a tested model, the low MAE and RMSE indicate the high prediction accuracy for missing data.

*B. Comparison with State-of-the-Art Models*

In the next, we compare the $\beta$-NLFT with three the state-of-the-art models, which are presented as follows.
1) M1: A HDOP model [33], which takes into account the multi-dimensional properties of client QoS and models multi-dimensional QoS data by a multi-linear generation approach.
2) M2: A TCA model [18], which draws on the reference reporting mechanism, comes out with a new method for assessing the QoS providers in a service-oriented environment.
3) M3: A BNLFT model [14], which incorporates linear biases and models non-negativity constraints into the model.
4) M4: The $\beta$-NLFT model proposed in this paper.

Fig. 2 show the RMSE and MAE of all tested models. From these results, we have: M4 is superior to its peers in term of prediction accuracy for unobserved dynamic QoS data. As shown in Fig. 2, M4 achieves the lowest RMSE and MAE in D1 and D2. For example, on D2, M4's RMSE is 18.474, which is about 40.26% lower than the 30.924 achieved by M1, 6.51% lower than the 19.761 achieved by M2, and 1.73% lower than the 18.8 achieved by M3. Considering the MAE on D2, as depicted in Fig.2 (d), the MAE of M4 is 3.125, which is 17.94%, 10.46% and 5.02% lower than M1's 3.808, M2's 3.49 and M3's 3.29, respectively. We can draw similar conclusions on D1, as shown in Fig. 2(a) and (b).

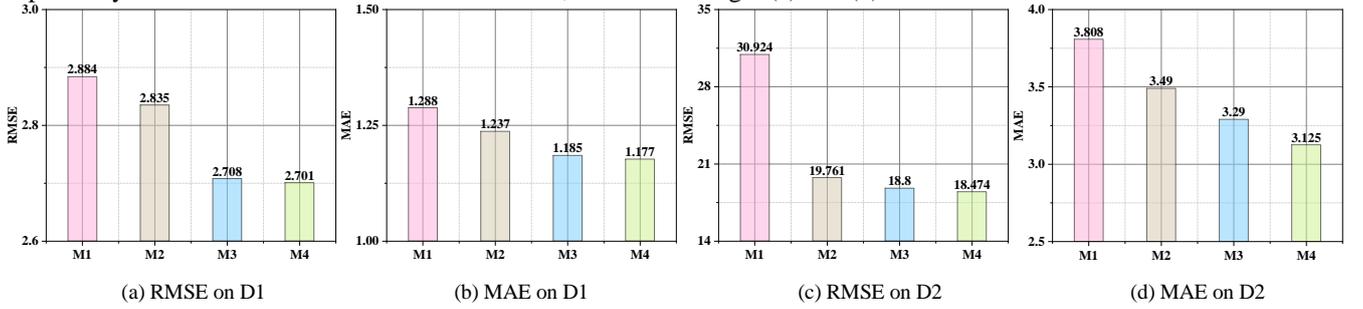

(a) RMSE on D1   (b) MAE on D1   (c) RMSE on D2   (d) MAE on D2

Fig. 2. The RMSE and MAE of M1-M4

## V. CONCLUSION

In this paper, we propose a time-aware QoS prediction model, i.e., $\beta$-NLFT model, for predicting unobserved dynamic QoS data. Compared with several state-of-the-art methods, it enjoys high prediction accuracy due to its $\beta$-divergence learning objective. Moreover, the performance of the $\beta$-NLFT model is validated on two dynamic QoS datasets. For the future, we intend to implement the model self-adaptation by using the latest evolutionary algorithms as well as extend the objective function to achieve better performance [35]-[39].